\documentclass{article}
\usepackage{ijcai19}

\usepackage{times}
\usepackage{soul}
\usepackage{url}
\usepackage[hidelinks]{hyperref}
\usepackage[utf8]{inputenc}
\usepackage[small]{caption}
\usepackage{booktabs}
\usepackage{cite}
\usepackage{amsmath,amssymb,amsfonts}
\usepackage{graphicx}
\usepackage{textcomp}
\usepackage{amsopn}
\usepackage[shortlabels]{enumitem}
\usepackage{etoolbox}
\usepackage{color}
\usepackage{fancyhdr, epsfig, epsf, subfigure}
\usepackage{threeparttable}
\usepackage{dsfont}
\usepackage{comment}
\usepackage{algorithm,algorithmic}
\usepackage{makecell}
\usepackage{booktabs, multirow}

\makeatletter
\let\@internalcite\cite
\def\cite{\def\citeauthoryear##1##2{##1, ##2}\@internalcite}
\def\shortcite{\def\citeauthoryear##1{##2}\@internalcite}
\def\@biblabel#1{\def\citeauthoryear##1##2{##1, ##2}[#1]\hfill}
\makeatother

\title{Federated Reinforcement Distillation with Proxy Experience Memory}

\author{
	Han Cha$^1$\footnote{Contact Author}\and
	Jihong Park$^2$\and
	Hyesung Kim$^1$\and
	Seong-Lyun Kim$^1$\And
	Mehdi Bennis$^2$\\
	\affiliations
	$^1$School of Electrical and Electronic Engineering, Yonsei University, Seoul, Korea\\
	$^2$Centre for Wireless Communications, University of Oulu, Finland\\
	\emails
	$^1$\{chan, hskim, slkim\}@ramo.yonsei.ac.kr,
	$^2$\{jihong.park, mehdi.bennis\}@oulu.fi
}

\begin{document}

\maketitle

	\begin{abstract}
		In distributed reinforcement learning, it is common to exchange the experience memory of each agent and thereby collectively train their local models. The experience memory, however, contains all the preceding state observations and their corresponding policies of the host agent, which may violate the privacy of the agent. To avoid this problem, in this work, we propose a privacy-preserving distributed reinforcement learning (RL) framework, termed \emph{federated reinforcement distillation (FRD)}. The key idea is to exchange a \emph{proxy experience memory} comprising a pre-arranged set of states and time-averaged policies, thereby preserving the privacy of actual experiences. Based on an advantage actor-critic RL architecture, we numerically evaluate the effectiveness of FRD and investigate how the performance of FRD is affected by the proxy memory structure and different memory exchanging rules.
	\end{abstract}

	\section{Introduction}
\begin{figure}[t]
			\subfigure[\small Policy distillation with experience  memory.]{\includegraphics[width=\linewidth]{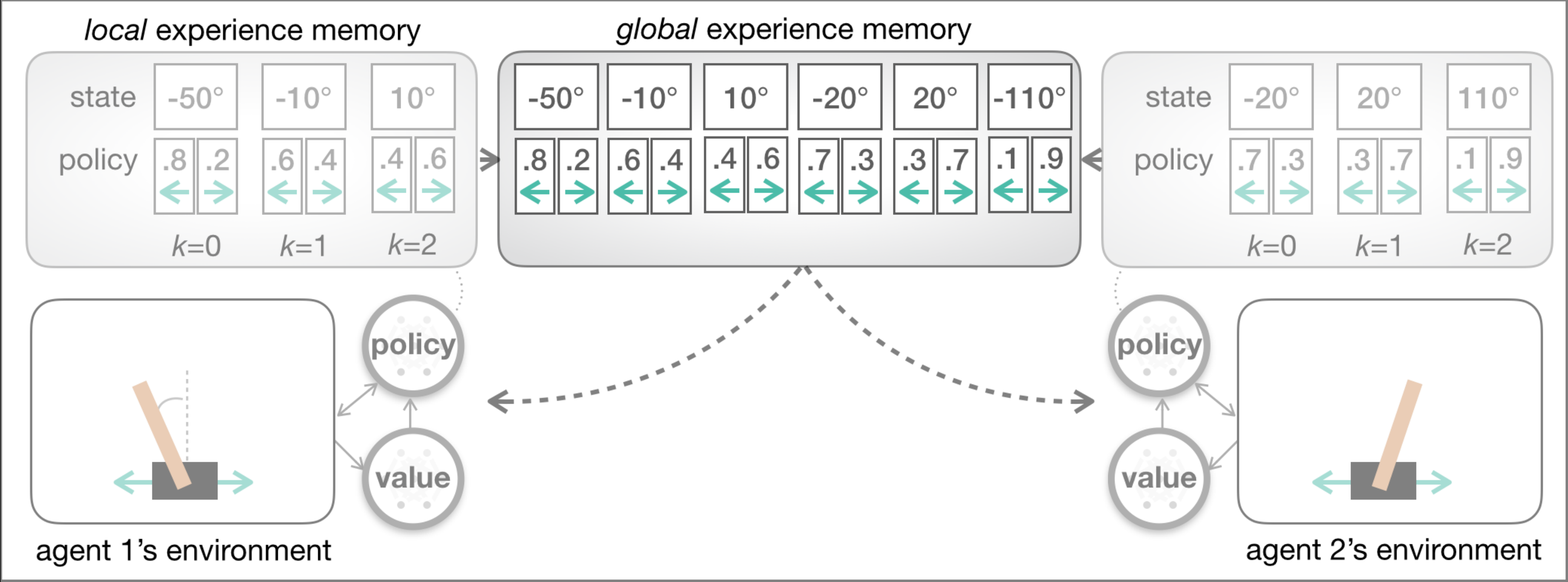}} 
			\subfigure[\small FRD with proxy experience memory.]{\includegraphics[width=\linewidth]{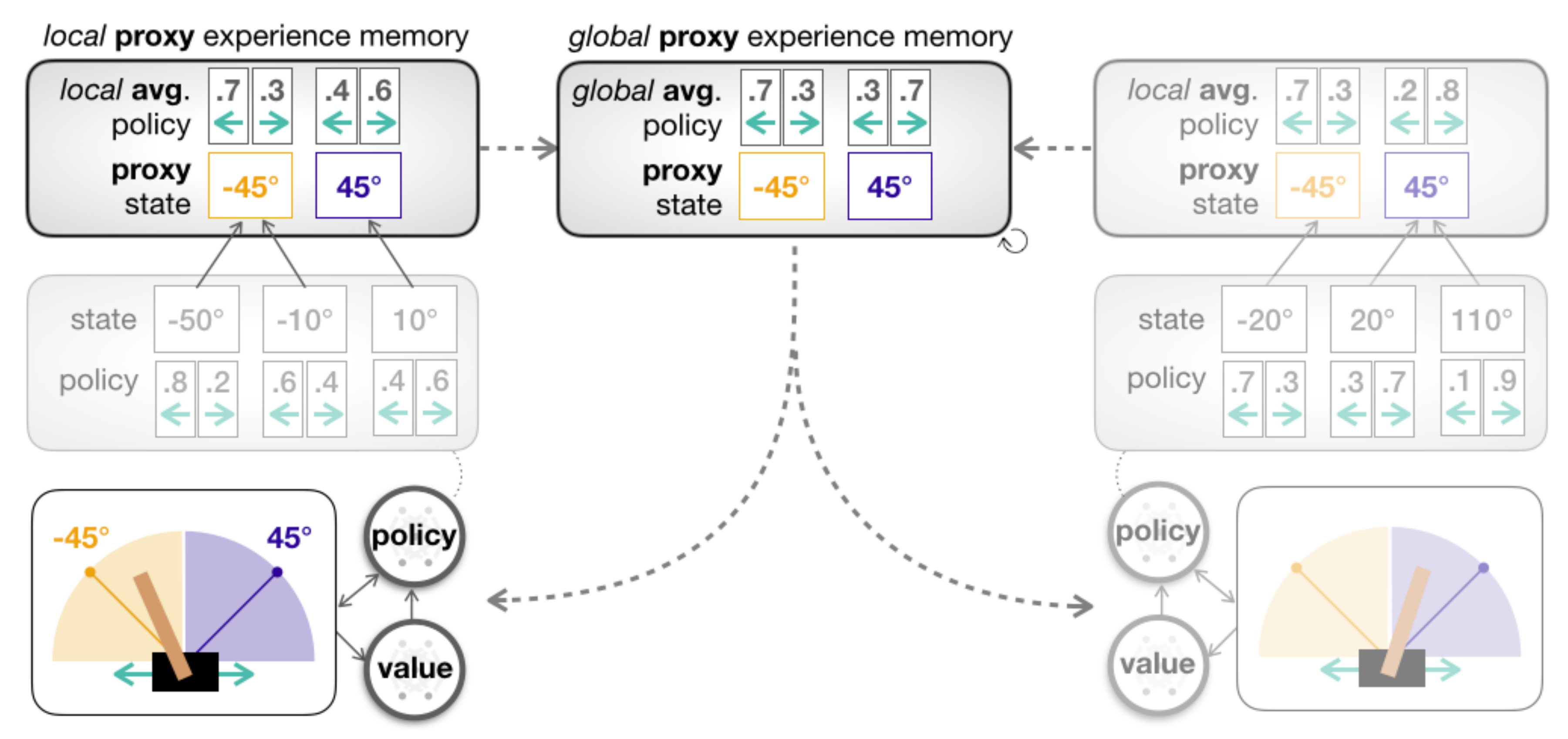}}
			\caption{\small Comparison between (a) a baseline distributed reinforcement learning (RL) framework, policy distillation with experience memory [Rusu \emph{et al.}, 2016], and (b) the proposed \textit{federated reinforcement distillation (FRD)} with \textit{proxy experience memory}.} \label{Fig:main_structure}
		\end{figure}

		Recent advances in mobile computing power have led to the emergence of intelligent autonomous systems \cite{Park:2018aa,Hamid:GC19}, ranging from driverless cars and drones to self-controlled robots in smart factories. Each agent therein interacts with its environment and carries out decision-making in real time. Distributed deep reinforcement learning (RL) is a compelling framework for such applications, in which multiple agents collectively train their local neural networks (NNs). As illustrated in Figure \ref{Fig:main_structure}(a), this is often done by: (i) uploading every local \emph{experience memory} to a server, (ii) constructing a global experience memory at the server, and (iii) downloading and replaying the global experience memory at each agent to train its local NN \cite{Rusu16}. However, a local experience memory contains all local state observations and the corresponding policies (i.e., action logits), and exchanging this may violate the privacy of its host agent.
		
		To obviate this problem, we propose a distributed RL framework based on a \emph{proxy experience memory}, which is termed \emph{federated reinforcement distillation (FRD)} and depicted in Figure \ref{Fig:main_structure}(b). In contrast to conventional experience memories containing actual states and policies, a local proxy experience memory at each agent consists of a set of pre-arranged \emph{proxy states} and \emph{locally averaged policies}. In this memory structure, the actual states are mapped into the proxy states, e.g., based on the nearest value rule, and the actual policies are averaged over time. Exchanging the local proxy memories of agents thereby preserves their privacy by hiding each agent's actual experiences from the others.
		
		In this work, we consider actor-critic RL architecture comprising two separate NNs, i.e., policy (actor) and value (critic) NNs, and study how to construct the local and global proxy memories, how to update each agent's local NN using the global proxy memory, and finally how the performance of the proposed FRD framework is changed along various proxy memory structure and different memory exchanging rules.
		
		\noindent\textbf{Related Works}.\quad
		Distributed deep RL has been investigated as policy distillation \cite{Rusu16} and advantage actor-critic (A2C) \cite{pmlr-v48-mniha16} algorithms, under policy NN and actor-critic based RL architectures, respectively. Both algorithms rely on exchanging actual experience memories. For classification tasks, distributed machine learning via exchanging NN outputs has been proposed as federated distillation (FD) in our preceding work \cite{Jeong18}. In FD, the outputs are quantized based on the classification labels, for maximizing communication efficiency. FRD leverage and extend this idea to distributed RL scenarios, in the context of its preserving privacy, rather than improving communication efficiency. It is noted that federated learning \cite{Brendan17} is another promising enabler for private distributed RL by exchanging NN model parameters, which has been recently studied as federated reinforcement learning (FRL) in \cite{Zhuo:2019aa}. Because of this, we conclude this paper by comparing FRL and our proposed FRD in the last section.
		
	\section{Background: Distributed Reinforcement Learning with Experience Memory}
		We consider the episodic, discrete state and action space Markov decision process, with state space $\mathcal{S}$, action space $\mathcal{A}$ and reward at each time slot denoted by $r_t \in \mathbb{R}$.
		The policy is stochastic and denoted by $\pi_\theta: \mathcal{S} \rightarrow \mathcal{P}(\mathcal{A})$, where $\mathcal{P}(\mathcal{A})$ is the set of probability measures on $\mathcal{A}$. 
		The parameters of local model are denoted by $\theta \in \mathbb{R}^n$, and $\pi_\theta(a|s)$ is the conditional probability of $a$ when the state is $s$.
		The reinforcement learning (RL) interacts with the environment without any prior knowledge about the environment.
		
		In policy distillation presented in \cite{Rusu16}, the agents $i=1,\cdots,U$ construct the dataset named \textit{experience memory}  for training local model $\theta_i$. The experience memory $\mathcal{M}=\{(s_k, \pi(\mathbf{a}_k|s_k))\}^{N}_{k=0}$ consists of the state $s_k$ and the policy vector $\pi(\mathbf{a}_k|s_k)$ tuple, where $\mathbf{a}=(a^1,\cdots,a^{|\mathcal{A}|})$ is action vector. As illustrated in Figure \ref{Fig:main_structure}(a), the experience memory $\mathcal{M}$ is collected with following procedures.
		\begin{itemize}
			\item Each agent records the \textit{local experience memory} $\mathcal{M}_i=\{(s_k, \pi_{\theta_i}(\mathbf{a}_k|s_k))\}^{N_i}_{k=0}$ tuple during $E$ episodes. The size of local experience replay $N_i$ is identical with the learning steps. In this paper, we assume that all the agents wait for the last agent completing the episode.
			\item After all the agents complete the $E$ episodes, the server collects $\mathcal{M}_i$ of each agent.
			\item Then, the server constructs a \textit{global experience memory} $\mathcal{M}=\{(s_k, \pi(\mathbf{a}_k|s_k))\}^{N}_{k=0}$, where $N=\sum_{i=1}^{U}N_i$ and $\pi$ denotes the arbitrary policy of agents.
		\end{itemize}
		
		After the global experience memory $\mathcal{M}$ is constructed, the agents update their local model $\theta_i$ with following procedures.
		\begin{itemize}
			\item To reflect the knowledge of other agents, the agents download the global experience memory $\mathcal{M}$ from the server.
			\item Similar to the conventional classification setting, each agent $i$ fits the local model $\theta_i$ minimizing the cross entropy loss $L_i(\mathcal{M},\theta_i)$ between the policy of local model $\pi_{\theta_i}(\mathbf{a}_k|s_k)$ and the policy $ \pi $ of global experience memories $\mathcal{M}$, where
			\begin{align}
			L_i(\mathcal{M}, \theta_i)=-\sum_{k=1}^{N}\pi(\mathbf{a}_k|s_k)\log\left(\pi_{\theta_i}(\mathbf{a}_k|s_k)\right).
			\end{align}
		\end{itemize}
		
		Unfortunately, direct exchanging the local experience memories of agents has privacy leakage issue. The server can get all information about the states visited by the host agents and the corresponding policy of the host agents. Because of that, privacy leakage issue is inevitable to utilize the policy distillation method.

	\section{Federated Reinforcement~Distillation (FRD) with Proxy Experience~Memory}
		In this section, we introduce the novel \textit{federated reinforcement distillation (FRD)} framework that provides a privacy-preserving communication-efficient federated reinforcement distillation. The agents utilizing the FRD construct the novel dataset named \textit{proxy experience memory} $\mathcal{M}^P=\{(s_k^p,\pi^p\left(\mathbf{a}_k|s_k^p\right)\}^{N^P}_{k=0}$, where the $s^p$ denotes the \textit{proxy state} and the $\pi^p\left(\mathbf{a}_k|s_k^p\right)$ denotes \textit{average policy}. The proxy state is representative state of \textit{state cluster} $C_j\in \mathcal{C}$. Note that the union of proxy state cluster sets is the state space $\mathcal{S}$, i.e., $\mathcal{S}=\bigcup_{j=1}^{|\mathcal{C}|}C_j$ and none of the state cluster has the joint set, i.e., $ C_i \cap C_j = \emptyset, i\neq j $.
		
		As illustrated in Figure \ref{Fig:main_structure}(b), the proxy experience memory $\mathcal{M}^P$ is formed with following procedures.
		\begin{itemize}
			\item Each agent categorizes the policy $\pi_{\theta_i}(\mathbf{a}|s)$ along the states $s$ included in the proxy state cluster, i.e., $s\in C_j$.
			\item After all the agents complete the $E$ episodes, each agent calculates \textit{local average policy} $ \pi_{\theta_i}^p\left(\mathbf{a}_k|s_k^p\right) $ by averaging the policy $\pi_{\theta_i}(\mathbf{a}|s)$ in the proxy state cluster $C_j$ and make \textit{local proxy experience memory} $\mathcal{M}^P_i=\{(s_k^p,\pi_{\theta_i}^p\left(\mathbf{a}_k|s_k^p\right)\}^{N_i^P}_{k=0}$. The size of local proxy experience memory $N_i^P$ is identical with the number of proxy state cluster that visited by the agent.
			\item When the local proxy experience memories of every agent is ready, the server collects $ \mathcal{M}^P_i $ of each agent.
			\item Then, the server constructs the \textit{global proxy experience memory} $\mathcal{M}^P=\{(s_k^p,\pi^p\left(\mathbf{a}_k|s_k^p\right)\}^{N^P}_{k=0}$ by averaging the local average policy of local proxy experience memory along the state cluster. The size of global proxy experience memory $ N^P $ is identical with the number of proxy state cluster that visited by all agents.
		\end{itemize}
		As same as the policy distillation case, the agents utilizing the FRD update their local model $\theta_i$ with following procedures.
		\begin{itemize}
			\item As the distributed RL procedure, the agents download the global proxy experience memory $ \mathcal{M}^P $ from the server.
			\item Each agent $ i $ fits the local model $ \theta_i $ minimizing the cross entropy loss $ L^P_i(\mathcal{M}^P,\theta_i) $ between the policy of local model $\pi_{\theta_i}(\mathbf{a}_k|s_k)$ and the global average policy $ \pi^p(s_k^p,\mathbf{a}_k|s_k^p) $ of global proxy experience memory $ \mathcal{M}^P $, where
			\begin{align}
			L_i^P(\mathcal{M}^P, \theta_i)=-\sum_{k=1}^{N^P}\pi^p(\mathbf{a}_k|s_k^p)\log\left(\pi_{\theta_i}(\mathbf{a}_k|s_k^p)\right).
			\end{align}
			We note that the loss is calculated with the policy produced by the local model as the input of proxy state.
		\end{itemize} 
		
		As we mentioned above, exchanging the proxy experience memories keeps the privacy of the host agents. The server merely can get about the information of proxy states that the host agents visited in the learning stages. With the same perspective, the server also only can get the average policy of host agents. As a result, the host agents protect their privacy by utilizing our proposed framework.
		
		Furthermore,  the size of global proxy experience memory is much smaller than that of experience memory due to state clustering. When the memory sharing occurs through a wireless channel, the payload size is a key factor of sharing feasibility. In this point of view, FRD provides a communication-efficient distributed RL framework. 
	
	\section{FRD under Actor-Critic Architectures}
	
		\begin{figure}[t]
			\centering
			\includegraphics[width=8cm]{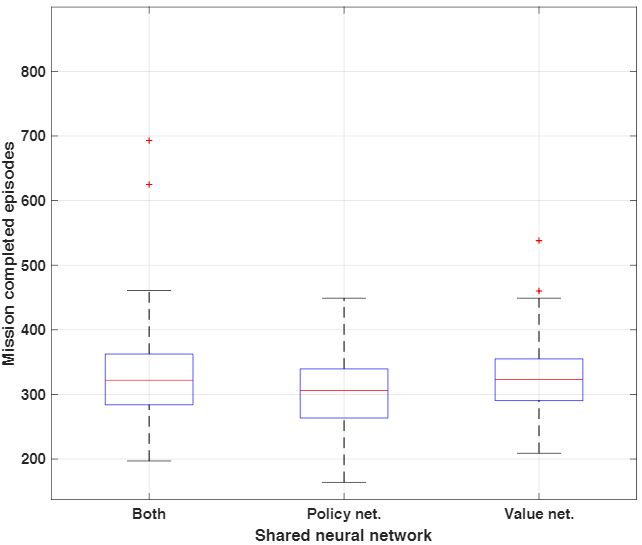}
			\caption{Performance comparison according to exchanging model. The case of exchanging policy network is better than other cases in terms of performance variation.} \label{Fig:PvsV}
		\end{figure}
		
		The advantage actor-critic (A2C) algorithm~\cite{pmlr-v48-mniha16} consists of two parts, actor and critic NNs. The actor generates the action $a \in \mathcal{A}$ according to the policy $\pi_\theta$, and the critic evaluates selected action how much is beneficial than other actions for gaining a more expected future reward. But the actor and the critic have no prior knowledge of the environment; the actor-critic pair have to interact with the environment and learn optimal policy to getting a maximum expected future reward. By adopting the neural network structure, the actor and the critic effectively learn optimal policy $ \pi^* $.
		
		The advantage function~\cite{Wang:2016aa} is the metric evaluating the action generated by the actor. If the value of the advantage function is positive, it means that the selected action is not the optimal compared to other actions. In other words, the advantage function $A$ is defined as follows:
		
		\begin{align} \label{eq:advantage}
			A^\pi(s_t,a_t)=& Q^\pi(s_t,a_t)-V^\pi(s_t) \\ \nonumber
			=& r(s_t,a_t) + \mathbb{E}_{s_{t+1}\sim\mathbb{E}}\left[V^\pi(s_{t+1})\right]-V^\pi(s_t) \\ \nonumber
			\approx& r(s_t,a_t) + V^\pi(s_{t+1}) - V^\pi(s_t).
		\end{align}
		where $ Q^\pi(s,a)=\mathbb{E}\left[r_0^\gamma|s_0=s,a_0=a;\pi\right] $, the value function $ V^{\pi}(s)=\mathbb{E}[r_0^\gamma|s_0=s;\pi]$, and $ r(s_t,a_t) $ is instant reward at learning step $ t $.
		As we can see in equations (\ref*{eq:advantage}), we can obtain the advantage function with just only value function. As a result, the neural network of the critic approximates the value function and estimates the advantage function in every updating step of the policy network.
		
		Under the A2C algorithm, we have to select which model to learn using the FRD framework - only one among two models or both? As we mentioned in section 2 and 3, the policy network forms the experience memory with the policy $ \pi $. Similarly, the value network creates the \textit{value memory} that consists of the state and corresponding value pairs. In the FRD case, the \textit{average value} replaces the average policy.
		
		In Figure \ref*{Fig:PvsV}, we represent the performance comparison in each case: both, policy network, and value network. Three cases have similar performance in terms of the number of episodes until complete the mission. Unlike two other cases, the case exchanging the policy network shows stable learning results, i.e., the variation of mission completion time is smaller than two other cases. For this reason, we select the policy network to apply FRD framework. In the rest of the paper, we utilize the FRD framework with the experience memory made by the output of the policy network unless we mention it.

	\section{Experiments}
		The group of the RL agents shares the output of the policy network to construct the proxy experience memory $ \mathcal{M}^P $ utilizing federated reinforcement distillation under the advantage actor-critic algorithm. In this paper, we implement the proposed federated reinforcement distillation framework in the \textit{Cartpole-v1} in \emph{OpenAI gym} environment to evaluating the performance. We evaluate the performance of propose FRD framework in terms of the number of episodes until the group of agent completes the mission. The mission of the group of agents is defined as achieving the average standing duration of the pole over ten episodes exceed the predetermined time duration. We assume that the group of agents complete the mission if just one of the agents in the group completes the mission. Each agent adopts the advantage actor-critic model for the local model and the model size of the policy network presented in Table \ref{tab:parameters}. Note that the model size of the value network is identical to that of the policy network.
		
		\begin{figure*}[t!]
			\centering
			\subfigure[Setting  1.] {\includegraphics[width=4.3cm]{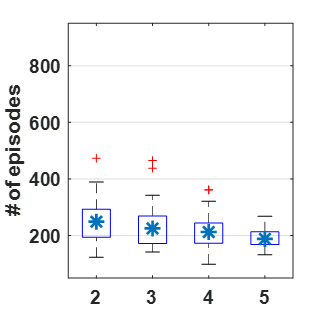}}
			\subfigure[Setting  2.] {\includegraphics[width=4.3cm]{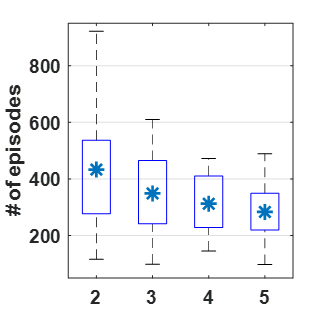}}
			\subfigure[Setting  3.] {\includegraphics[width=4.3cm]{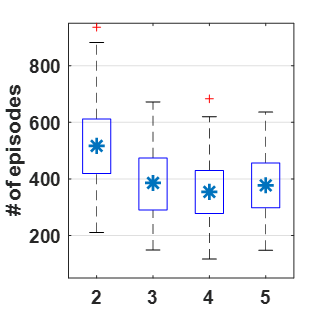}}
			\subfigure[Setting  4.] {\includegraphics[width=4.3cm]{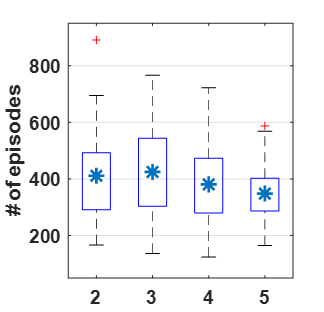}}
			\subfigure[Setting  5.] {\includegraphics[width=4.3cm]{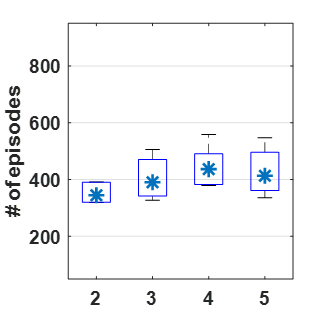}}
			\subfigure[Setting  6.] {\includegraphics[width=4.3cm]{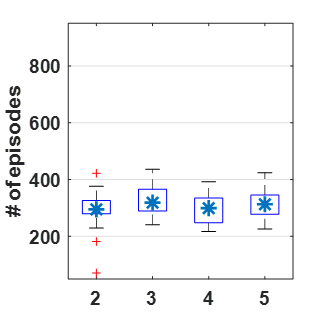}}
			\subfigure[Setting  7.] {\includegraphics[width=4.3cm]{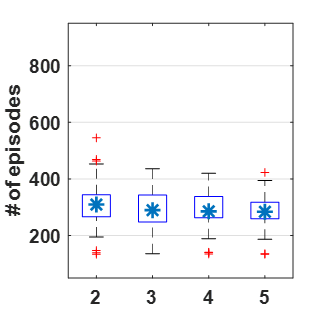}}
			\setlength{\belowcaptionskip}{-10pt}
			\setlength{\abovecaptionskip}{-2pt}
			\caption{Simulation results in \emph{Cartpole} environment. The x-axis label of all graphs is the number of agents in a cooperative group, and the y-axis label of all graphs is the number of episodes until the agent group completes the mission. The mission of the group of agents is defined as achieving the average standing duration of ten consecutive episodes exceeds the predetermined time duration. We assume that the group of agents completes the mission if one of the agents in the group completes the mission. The agents make the proxy experience memory with the output of the policy network only.} \label{Fig:main_results}
		\end{figure*}
		
		Before implement the federate reinforcement distillation, pre-arranging of the state clustering is needed. The state of \textit{Cartpole} environment consists of four component, which is position of cart, velocity of cart, angle of pole, velocity of pole tip. We evenly divide the each component with the number as $ S $ subsections. Then, we form the state cluster as the combination of divided components. As a result, the number of state cluster $ |\mathcal{C}| $ is identical with $ S^4 $. The proxy state of each state cluster is defined as the middle value of each subsection of components. For example, the proxy state of the corresponding state cluster $ C_j $ is $ s_p = [0.5, -0.75, 0.75, 0.05] $ when $ C_j = \{[0,1), [-1,-0.5), [0.5, 1), [0, 0.1)\} $.
		
		We perform the simulations with various hyperparameter settings presented in Table \ref{tab:parameters}, and corresponding results are presented in Figure \ref{Fig:main_results}. We investigate the impact of each hyperparameter in terms of the performance of each group. The box in Figure \ref{Fig:main_results} represents the data from 25\% to 75\%. The blue star represents the average of data. The red line represents the median of data.
		
		\begin{table}
			\centering
			\caption{Hyperparameters of federated reinforcement distillation.}
			\label{tab:parameters}
			\resizebox{
				0.48\textwidth}{!}{
				\begin{tabular}{c|c|c|c|c|c}
					\toprule
					\thead{Setting } & \thead{ \# of proxy \\ states ($S^4$)} & \thead{Memory exchange \\ period ($ E $)} & \thead{Initial learning \\ time ($ I $)} & \thead{ \# of weights per \\ hidden layer ($ n $)} & \thead{ \# of \\  hidden layers}\\
					\midrule
					\textit{1} & $100^4$ & 25 & 50 & \textbf{24} & 2 \\
					\textit{2} & $\mathbf{100^4}$ & 25 & \textbf{50} & \textbf{100} & 2 \\
					\textit{3} & $100^4$ & 25 & \textbf{100} & 100 & 2 \\
					\textit{4} & {$\mathbf{50^4}$} & 25 & 50 & 100 & 2 \\
					\textit{5} & $100^4$ & \textbf{10} & \textbf{0} & 24 & 1 \\
					\textit{6} & $100^4$ & \textbf{50} & \textbf{0} & 24 & 1 \\
					\textit{7} & $100^4$ & 25 & \textbf{125} & 24 & 1\\	
					\bottomrule          
				\end{tabular}
			}
		\end{table}
		
		\vspace{5pt}\noindent \textbf{Impact of the Proxy State Size}.\quad In the \emph{Setting 2} and \emph{Setting 4}, we can observe the impact of the proxy state size on the performance of FRD. When the multiple agents cooperate, the performance of \emph{Setting 4} is better than that of \emph{Setting 2} in terms of the average number and the variance of episodes. As the number of agents is increasing, the relation is reversed. Because the policy resolution of proxy state with smaller size is low, the knowledge of agents is blurred compare to that of proxy state with a bigger size. 
		
		Nevertheless, multiple agents case of \emph{Setting 4} has better performance though the proxy state size is 16 times smaller than that of \emph{Setting 2}. It means that the group of agents choose the proxy state size to reduce the payload size of exchanging information. If the agents cooperate through a wireless channel, they can select the proper state cluster size sacrificing a bit of learning performance. 
		
		\vspace{5pt}\noindent \textbf{Impact of Memory Exchange Period}.\quad In \emph{Case 5}, the performance of the group of agents is getting worse as the number of agents is increasing. Too frequent memory exchange and the local model update have no merit in increasing the number of agents. As shown in the \emph{Setting 6}, a moderate frequency of memory exchange brings stable performance enhancement.
		
		\vspace{5pt}\noindent \textbf{Impact of Initial Learning Time}.\quad If there is no initial learning time before exchanging the experience memory, the performance of FRD is degraded as well as unstable. In the \emph{Setting 5}, the absence of initial learning time results in performance degradation as the number of agents is increasing. The local model of the agent is not trained enough to exchange their proxy experience memory. Furthermore, too long initial learning time is also negative to the performance of FRD. Because too long initial learning time may give a chance of learning the bad policy of the individual local model of the agent, the cooperation of agents is getting worse the training of the local model of each agent. Comparing the \emph{Setting 2} and \emph{Setting 3}, the performance of the \emph{Setting 2} is better than that of the \emph{Setting 3}. As a result, the initial learning time should be selected properly to achieve higher performance.

		\vspace{5pt}\noindent \textbf{Impact of Neural Network Model Size}.\quad As we can see in \emph{Setting 1} and \emph{Setting 2}, smaller NN has better performance in terms of the number of episodes until the group agent completes the mission. Because we measure how fast the group agent completes the mission, bigger NN has a disadvantage in terms of convergence duration. In future work, the advantage of big NN compare to small NN can be evaluated in the more complex and score-pursuing environment like \emph{Atari games} in \emph{OpenAI gym}. On the other hand, too small NN has marginal gain about FRD. In \emph{Setting 6} and \emph{Setting 7}, the performance enhancement along increment of the number of agents is limited in certain average value boundary.

	\section{Discussion and Concluding Remarks}
		In this paper, we introduce a privacy-preserving communication-efficient distributed reinforcement learning framework, termed \emph{federated reinforcement distillation (FRD)}. The key idea is to exchange a \emph{proxy experience memory} comprising a pre-arranged set of states and time-averaged policies. It conceals the actual visited states and actions of host agents and additionally has the benefit of reduced memory size. In a communication-constrained situation, e.g., through a wireless channel, the proposed FRD framework has the advantage of existing policy distillation.
		
		Based on advantage actor-critic (A2C) algorithm, we evaluate the performance of FRD in various proxy memory structure and different memory exchanging rules. First, we investigate the impact of proxy memory structure on which network is used for FRD in the A2C algorithm - policy network, value network, or both. Second, based on the first investigation, we implement policy network based FRD and evaluate the performance in the various settings of memory exchanging rules - when, how often, how large the memory size it is, and how large the neural network size it is.
		
		As the future work, a performance comparison between federated learning and the FRD is promising. We evaluate the performance in a simple setting when multiple agents collaborate. The performance in terms of the average number of episodes until the group of the agent completes the mission is reletively equivalent. But in terms of variation, the proposed FRD has better performance than federated learning. The performance difference is due to the amount of noise when knowledge transfer occurs. It means that the noise of FRD is less than that of federated learning. The proposed FRD framework may be more suitable than the federated learning for the application that has a conservative constraint for performance degradation.
		
	\section*{Acknowledgements}
		This work was supported partly by Institute for Information \& communications Technology Promotion (IITP) grant funded by the Korea government (MSIT) (No. 2018-0-00923, Scalable Spectrum Sensing for Beyond 5G Communication) and Academy of Finland projects SMARTER, CARMA, and 6Genesis Flagship (grant no. 318927), and partly by AIMS and ELLIS projects at the University of Oulu.
	
		\begin{figure}[t]
			\centering\includegraphics[width=8cm]{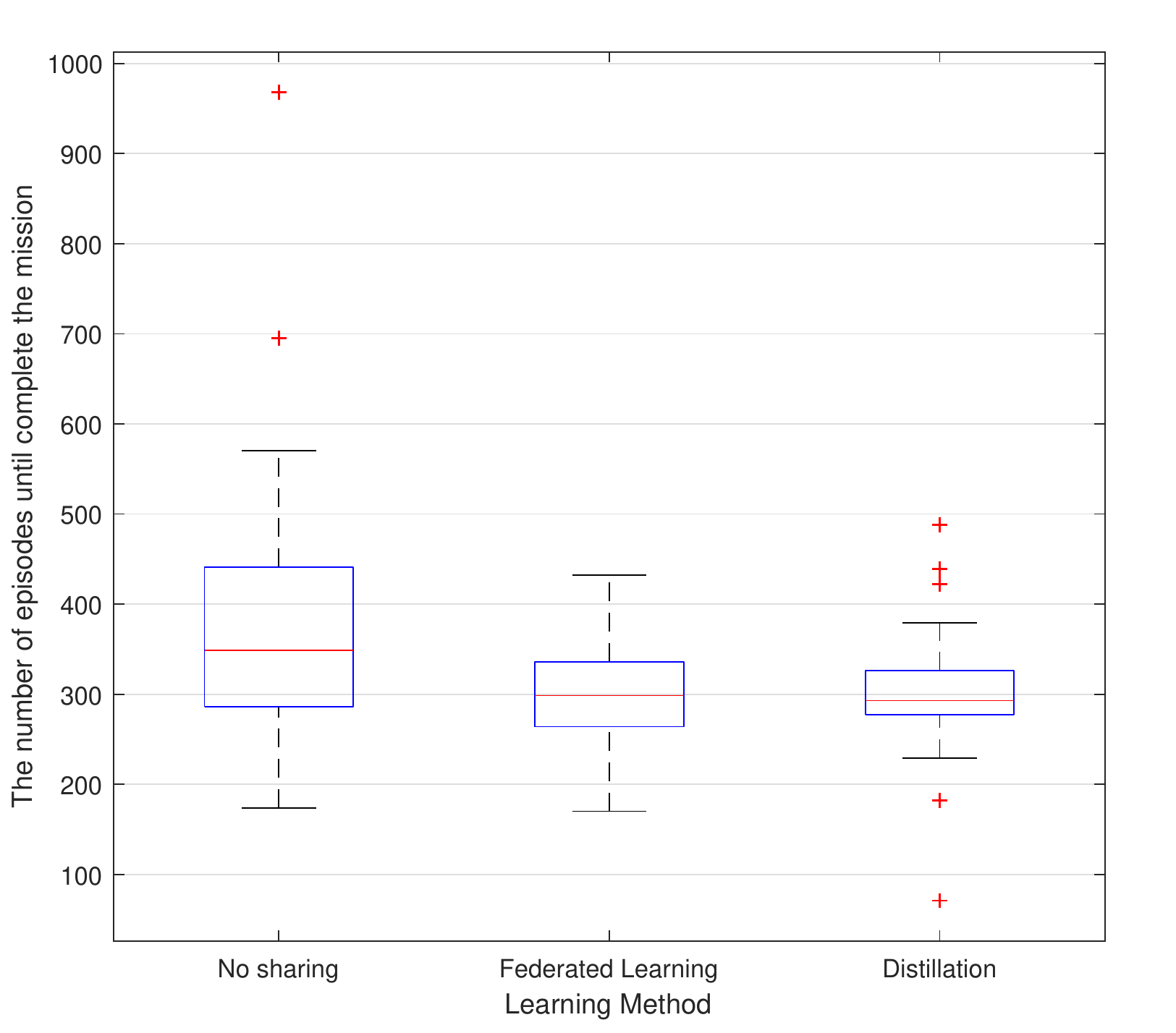}
			\setlength{\belowcaptionskip}{-10pt}
			\setlength{\abovecaptionskip}{-0.5pt}
			\caption{Performance comparison between federated learning and federated reinforcement distillation when the multiple agents collaborate.} \label{Fig:fvsd}
		\end{figure}

	\bibliographystyle{named}
	\bibliography{ijcai19}

\begin{thebibliography}{}

\bibitem[\protect\citeauthoryear{Jeong \bgroup \em et al.\egroup
  }{2018}]{Jeong18}
Eunjeong Jeong, Seungeun Oh, Hyesung Kim, Jihong Park, Mehdi Bennis, and
  Seong-Lyun Kim.
\newblock {Communication-efficient on-device machine learning: Federated
  distillation and augmentation under non-IID private data}.
\newblock In {\em Proc. {NeurIPS Workshop on Machine Learning on the Phone and
  other Consumer Devices (MLPCD)}}. Montr{\'e}al, Canada, December 2018.

\bibitem[\protect\citeauthoryear{McMahan \bgroup \em et al.\egroup
  }{2017}]{Brendan17}
H.~B. McMahan, E.~Moore, D.~Ramage, S.~Hampson, and B.~A. y~Arcas.
\newblock Communication-efficient learning of deep networks from decentralized
  data.
\newblock In {\em Proc. of AISTATS}, Fort Lauderdale, FL, USA, April 2017.

\bibitem[\protect\citeauthoryear{Mnih \bgroup \em et al.\egroup
  }{2016}]{pmlr-v48-mniha16}
Volodymyr Mnih, Adria~Puigdomenech Badia, Mehdi Mirza, Alex Graves, Timothy
  Lillicrap, Tim Harley, David Silver, and Koray Kavukcuoglu.
\newblock Asynchronous methods for deep reinforcement learning.
\newblock In Maria~Florina Balcan and Kilian~Q. Weinberger, editors, {\em
  Proceedings of The 33rd International Conference on Machine Learning},
  volume~48 of {\em Proceedings of Machine Learning Research}, pages
  1928--1937, New York, USA, 20--22 Jun 2016. PMLR.

\bibitem[\protect\citeauthoryear{Park \bgroup \em et al.\egroup
  }{2018}]{Park:2018aa}
Jihong Park, Sumudu Samarakoon, Mehdi Bennis, and M\'{e}rouane Debbah.
\newblock Wireless network intelligence at the edge.
\newblock {\em submitted to Proc. IEEE \emph{[Online]. ArXiv preprint:
  https://arxiv.org/abs/1812.02858}}, 2018.

\bibitem[\protect\citeauthoryear{Rusu \bgroup \em et al.\egroup
  }{2016}]{Rusu16}
A.~Rusu, S.~Colmenarejo, C.~Gulcehre, G.~Desjardins, J.~Kirkpatrick, and
  R.~Pascanu.
\newblock Policy distillation.
\newblock {\em ICRL}, 2016.

\bibitem[\protect\citeauthoryear{Shiri \bgroup \em et al.\egroup
  }{2019}]{Hamid:GC19}
Hamid Shiri, Jihong Park, and Mehdi Bennis.
\newblock Massive autonomous {UAV} path planning: A neural network based
  mean-field game theoretic approach.
\newblock {\em submitted to GLOBECOM 2019 \emph{[Online]}. ArXiv preprint:
  https://arxiv.org/abs/1905.04152}, 2019.

\bibitem[\protect\citeauthoryear{Wang \bgroup \em et al.\egroup
  }{2016}]{Wang:2016aa}
Ziyu Wang, Tom Schaul, Matteo Hessel, Hado Van~Hasselt, Marc Lanctot, and Nando
  De~Freitas.
\newblock Dueling network architectures for deep reinforcement learning.
\newblock {\em Proceedings of the 33rd International Conference on
  International Conference on Machine Learning}, 48:1995--2003, 2016.

\bibitem[\protect\citeauthoryear{Zhuo \bgroup \em et al.\egroup
  }{2019}]{Zhuo:2019aa}
Hankz~Hankui Zhuo, Wenfeng Feng, Qian Xu, Qiang Yang, and Yufeng Lin.
\newblock Federated reinforcement learning.
\newblock {\em \emph{[Online]}. Arxiv preprint:
  https://arxiv.org/abs/1901.08277}, 2019.

\end{thebibliography}

\end{document}